\documentclass[10pt,twocolumn,twoside]{IEEEtran}

\RequirePackage{snapshot}

\usepackage{caption}
\usepackage{subcaption}
\usepackage{authblk}

\usepackage{moreverb}
\usepackage{verbatim}

\usepackage{acronym}
\usepackage{enumitem}
\usepackage{color}

\usepackage{parskip}% http://ctan.org/pkg/parskip

\usepackage{latexsym}
\usepackage{tabularx}
\usepackage{graphicx}
\usepackage{cite} % Conflicts with ICST style
\usepackage{setspace}
\usepackage{inconsolata}
\newcommand{\weblink}[1]{\texttt{#1}}

\usepackage{siunitx}
\usepackage[usenames,dvipsnames,table]{xcolor}

\usepackage{amsmath}
\usepackage{amsfonts}
\usepackage{amssymb}
\usepackage{amsthm}

\usepackage{algorithmic}
\usepackage{algorithm}
\usepackage{cases}

\usepackage[table]{xcolor}    % loads also »colortbl«
\usepackage[utf8]{inputenc}
\usepackage{authblk}
\usepackage{footmisc}

\PassOptionsToPackage{hyphens}{url}\usepackage{hyperref}

\renewcommand{\eqref}[1]{(\ref{eq:#1})}
\newcommand{\secref}[1]{\S\ref{sec:#1}}
\newcommand{\figref}[1]{Fig.~\ref{fig:#1}}
\newcommand{\tabref}[1]{Table~\ref{tab:#1}}

\newcommand{\ea}{\emph{et al.}}

\graphicspath{{images/}}

\begin{document}

\title{Detecting Road Surface Wetness from Audio: \\A Deep Learning Approach}

\author{Irman Abdi\'c, Lex Fridman, Erik Marchi,~\IEEEmembership{Student Member,~IEEE}, Daniel E. Brown, William Angell,
  Bryan~Reimer, Bj\"{o}rn Schuller,~\IEEEmembership{Senior Member,~IEEE}

% \thanks{Manuscript received Month 00, 201x; revised Month 00, 201x;
%     accepted Month 00, 201x. Support for this work was provided by the New England University Transportation Center, and the Toyota Class Action Settlement Safety Research and Education Program. The views and conclusions being expressed are those of the authors, and have not been sponsored, approved, or endorsed by Toyota or plaintiffs’ class counsel. The associate
%     editor coordinating the review of this manuscript and approving it
%     for publication was Name Surname.}

\thanks{Support for this work was provided by the New England University Transportation Center, and the Toyota Class Action Settlement Safety Research and Education Program. The views and conclusions being expressed are those of the authors, and have not been sponsored, approved, or endorsed by Toyota or plaintiffs' class counsel.}

\thanks{ I.~Abdi\'c, L.~Fridman, D.~E.~Brown, W.~Angell and B.~Reimer are with the 
    MIT AgeLab, CTL, Massachusetts Institute of Technology, Cambridge,
    MA, United States. E-mail: {\{abdic, fridman, danbr,
      wha, reimer\}}@mit.edu }% <-this % stops a space
      
\thanks{I.~Abdi\'c is also with the Department of Informatics, Technische Universit\"at M\"unchen, Munich, Germany. E-mail:
    irman.abdic@tum.de}% <-this % stops a space
    
\thanks{ E.~Marchi and B.~Schuller are with the 
    Machine Intelligence \& Signal Processing Group, MMK, Technische Universit\"at M\"unchen, Munich, Germany. E-mail: {\{erik.marchi, schuller\}}@tum.de }% <-this % stops a space
    
\thanks{B. Schuller is also with the Department of Computing,
    Imperial College London, United Kingdom, and Chair of Complex \& Intelligent Systems, University of Passau, Passau, Germany. E-mail: schuller@ieee.org}% <-this % stops a space
    
  % \thanks{Color versions of one or more of the figures in this paper
  %   are available online at http://ieeexplore.ieee.org.}
%  \thanks{Digital Object Identifier xx.xxxx/xxxx}
}

% \markboth{IEEE SIGNAL PROCESSING LETTERS, VOL. XX, NO. XX, MONTH
%   YEAR}%
% {Abdi\'c \MakeLowercase{\textit{et al.}}: Detecting Road Surface Wetness from Audio: A Deep Learning Approach}

\markboth{}{Abdi\'c \MakeLowercase{\textit{et al.}}: Detecting Road Surface Wetness from Audio: A Deep Learning Approach}

\maketitle%

\begin{abstract}%
  We introduce a recurrent neural network architecture for automated road surface wetness detection from audio of tire-surface interaction. The robustness of our approach is evaluated on 785,826 bins of audio that span an extensive range of vehicle speeds, noises from the environment, road surface types, and pavement conditions including international roughness index (IRI) values from 25\,in/mi to 1400\,in/mi. The training and evaluation of the model are performed on different roads to minimize the impact of environmental and other external factors on the accuracy of the classification. We achieve an unweighted average recall (UAR) of 93.2\,\% across all vehicle speeds including 0\,mph. The classifier still works at 0\,mph because the discriminating signal is present in the sound of other vehicles driving by. %Removing audio segments at speeds below 2.9\,mph from consideration improves the UAR to 100\,\%. In the case when the vehicle speed is below 2.9\,mph, we were able to discriminate between wet and dry road surfaces from ambient noises and achieve 74.5\,\% UAR. 
\end{abstract}


\section{Introduction}\label{sec:introduction}
\begin{figure}[h!]
  \centering
  \begin{subfigure}[b]{\columnwidth}\includegraphics[width=\columnwidth]{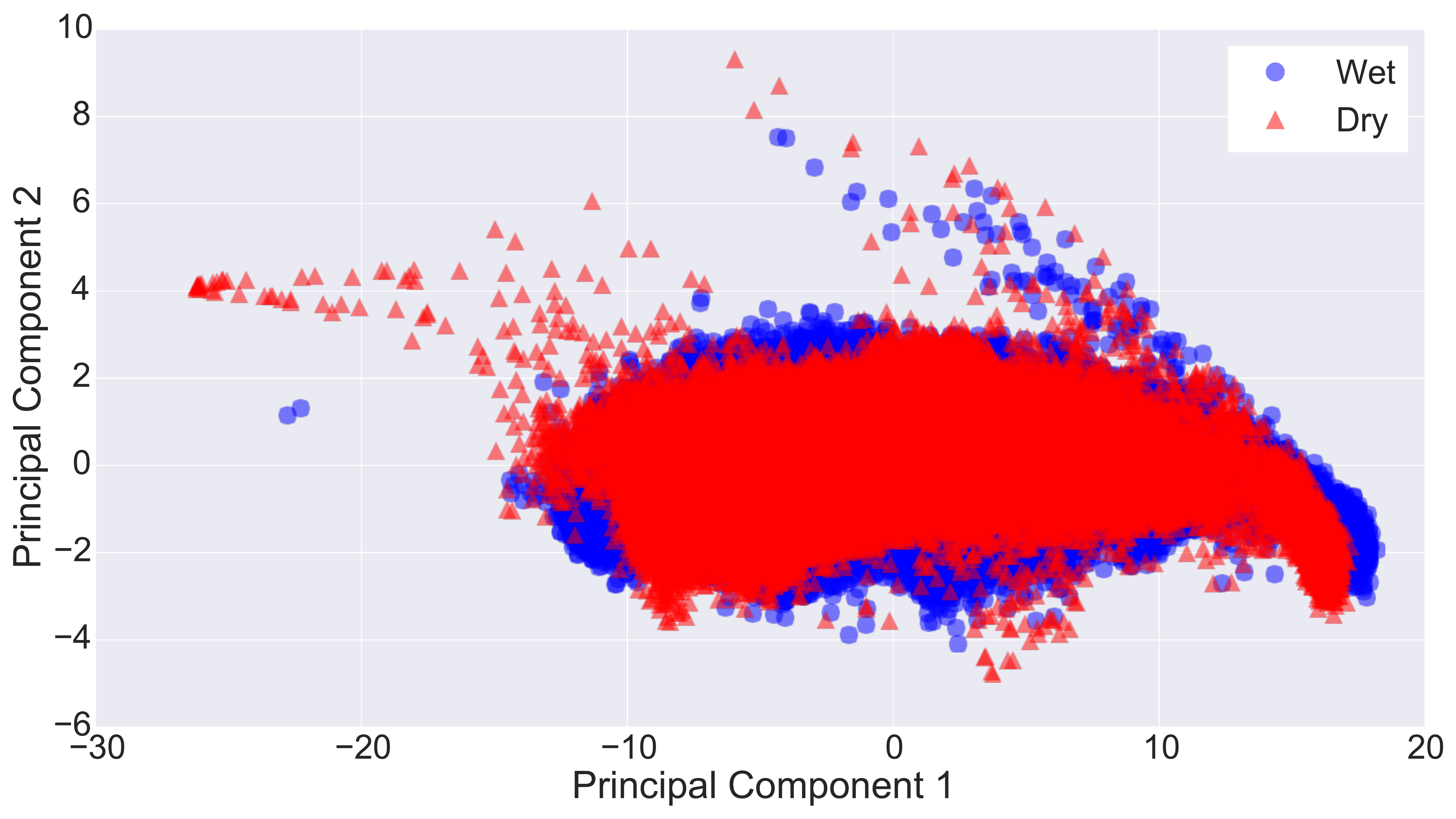}
    \caption{PCA for selected full wet and dry trips from \tabref{data-summary}.}
    \label{fig:pca-all}
  \end{subfigure}\\\vspace{0.1in}
  \begin{subfigure}[b]{\columnwidth}\includegraphics[width=\columnwidth]{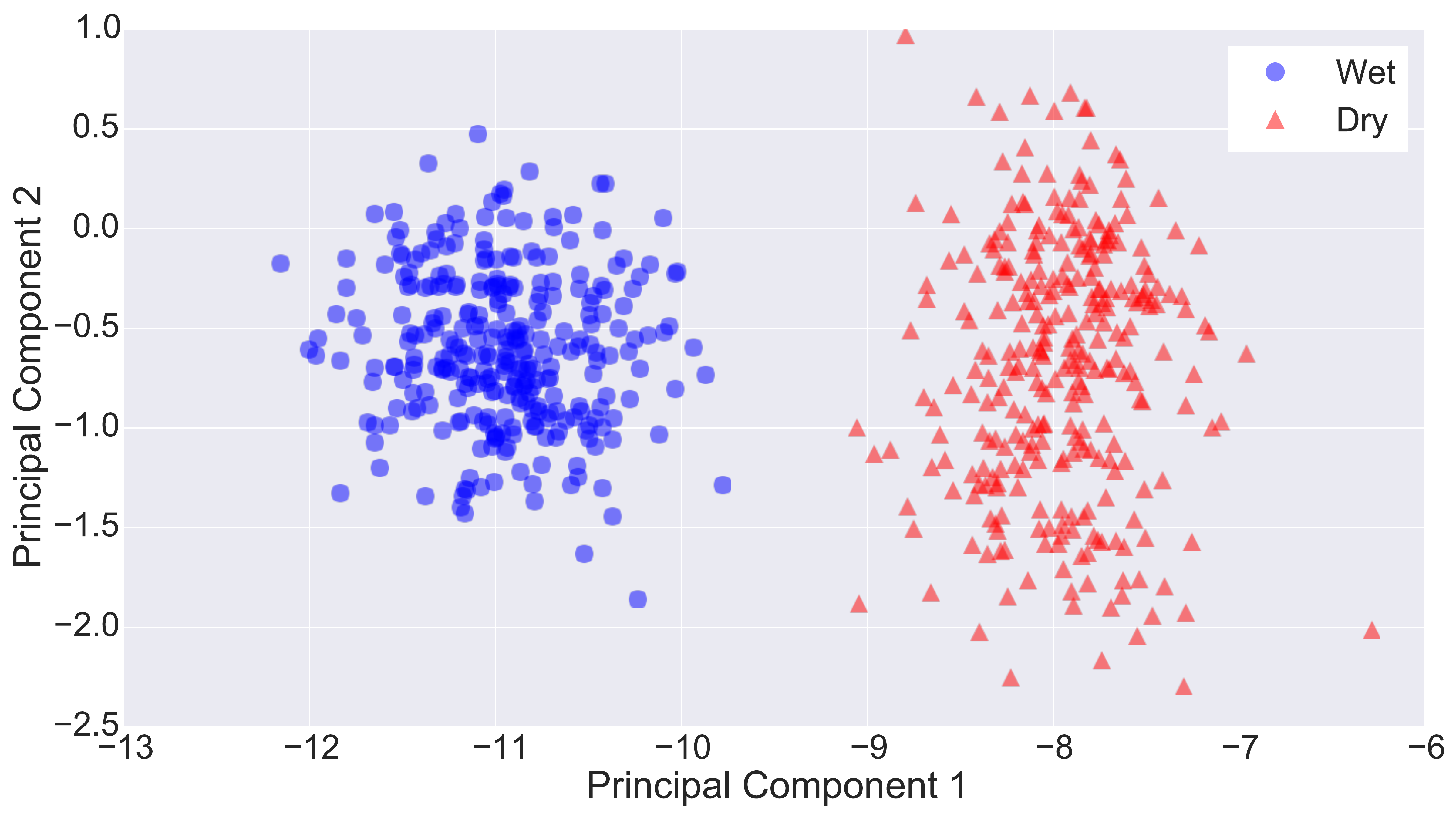}
    \caption{PCA for a randomly-selected segment of road from wet and dry trips from \tabref{data-summary}.}
    \label{fig:pca-segment}
  \end{subfigure}%
  \caption{PCA analysis for wet and dry road surfaces that illustrates a representative case where audio-based wetness
    detection is linearly separable for similar road type and vehicle speeds.}
  \label{fig:pca}
  \vspace{-2em}
\end{figure} 
Wet pavement is responsible for 74\,\% of all weather-related crashes in the U.S. with over 380,000 injuries and 4,700 deaths per year \cite{hamilton2012weather}. Furthermore, wet roads often increase traffic congestion and result in infrastructure damage and supply chain disruptions \cite{andrey2003weather}. From the perspective of driver safety, wetness detection during the period of time after the percipitation has ended but whether the road is still wet is critical. Under these conditions, human estimation of road wetness and friction properties is less accurate than normal, especially in reduced visibility over night or in the presence of fog \cite{andrey2001weather}.

The automated detection of road conditions from audio may be an important component of next generation Advanced Driver Assistance Systems (ADAS) that have the potential to enhance driver safety \cite{mueller2015sensor}. Moreover, autonomous and semi-autonomous vehicles have to be aware of road conditions to automatically adapt vehicle speed while entering the curve or keep a safe distance to the vehicle in front. There are numerous approaches that can detect whether a surface is wet or dry, but in the majority of cases they are not robust to variation in real-world datasets. Accuracy of video-based wetness prediction decreases significantly in poor lighting conditions (i.e., night, fog, smoke). Audio-based wetness prediction is heavily dependent upon surface type and vehicle speed which is fairly represented in our dataset of 785,826 bins (feature vectors described in \secref{audio-features}) \cite{alonso2014board}. We elucidate this dependence by visualizing the first two principal components for (1) two full trips and (2) a small 10-second section of road from (1). These two visualizations are shown in \figref{pca-all} and \figref{pca-segment}, respectively. The feature set we use is linearly separable for a specific road type and vehicle speed, as visualized in \figref{pca-segment}.
However, given the nonlinear relation of our feature set for (1) that is visualized in \figref{pca-all} we applied Recurrent Neural Networks (RNNs) which can model and separate the data points.

\section{Related Work}\label{sec:related-work}

Long short-term memory RNNs (LSTM-RNNs) have been successfully applied in many fields from hand writing recognition to robotic heart surgery \cite{graves2009novel,mayer2008system}. In the audio context, LSTM-RNNs contributed to the development of better phoneme classification, speech enhancement, affect recognition from speech, animal species identification and finding temporal structure in music \cite{graves2005framewise,wollmer2008abandoning,xu2014experimental,weninger2011audio,eck2002finding,Marchi14-MLP}. However, to our best knowledge LSTM-RNNs have not been applied to the task of road wetness detection.

Related works can be found in the video processing domain, where wetness detection has been studied with two camera set-ups: (1) a surveillance camera at night, and (2) a camera on-board a vehicle. The detection of road surface wetness using surveillance camera images at night is relying on passing cars' headlights as a lighting source that creates a reflection artifact on the road area \cite{horita2012efficient}. On-board video cameras use polarization changes of reflections on road surfaces or spatio-temporal reflection models \cite{yamada2003study,jokela2009road,amthor2015road}. 
A recent study uses near infrared (NIR) camera to classify several road conditions per every pixel with a high accuracy, the evaluation has been done in laboratory conditions, and field experiments  \cite{jonsson2015road}. However, a drawback of video processing methods is that they require (1) an external illumination source to be present and (2) visibility conditions to be clear.

Another approach capable of detecting road wetness relies on 24-GHz automotive radar technology for detecting low-friction spots \cite{viikari2009road}. It analyzes backscattering properties of wet, dry, and icy asphalt in laboratory and field experiments.

Traditionally, audio analysis of the road-tire interaction has been done by examining tire noises of passing vehicles from a stationary microphone positioned on the side of the road. This kind of analysis reveals that tire speed, vertical tire load, inflation pressure and driving torque are primary contributors to tire sound in dry road conditions \cite{iwao1996study}. Acoustic-based vehicle detection methods, as the one that uses bispectral entropy have been applied in the ground surveillance systems \cite{bao2009acoustical}. Other on-road audio collecting devices for surface analysis can be found in specialized vehicles for pavement quality evaluation (e.g., VOTERS \cite{birken2012voters}) and for vehicles instrumented for studying driver behavior in the context of automation (e.g., MIT RIDER \cite{fridman2015automated}). Finally, road wetness has been studied from on-board audio of tire-surface interaction, where SVMs have been applied \cite{alonso2014board}. 
\subsection{Contribution}
The method described in our paper improves the prediction accuracy of the method presented in \cite{alonso2014board} and expands the evaluation to a wider range of surface types and pavement conditions. Additionally, the present study is the first in applying LSTM-RNNs in this field. Moreover, we improve on the following three aspects of \cite{alonso2014board} where (1) the model was trained and tested on the same road segment, (2) false predictions caused by the impact of pebbles on the vehicle chassis were ignored, and (3) audio segments associated with speeds below 18.6 mph were removed.

We trained and tested the model on different routes, and considered all predictions regardless of the speed, pebbles impact or any other factor.

\section{Road Surface Wetness Classification}\label{sec:classification}

\subsection{Data Collection}\label{sec:data-collection}
\begin{figure}[h!]
  \centering
    \includegraphics[width=.24\textwidth]{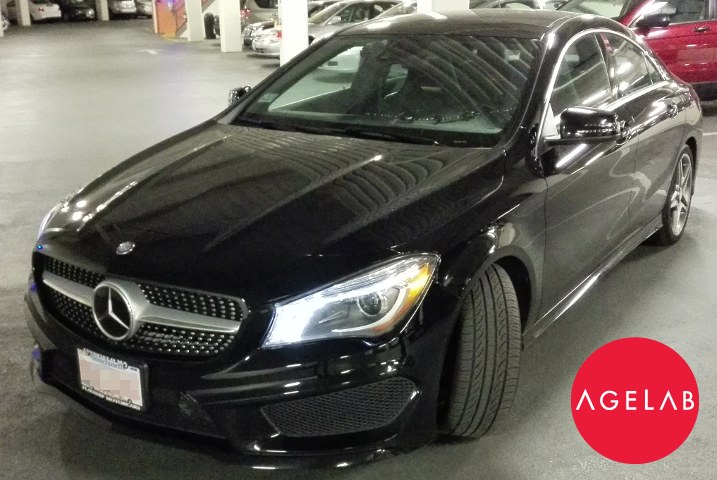}
    \includegraphics[width=.24\textwidth]{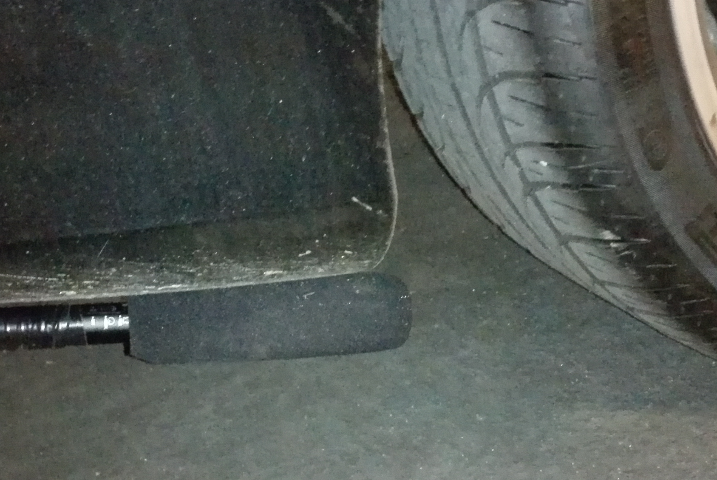}
    \caption{Instrumented MIT AgeLab vehicle (left) and placement of the shotgun microphone behind the rear tire (right).}
    \label{fig:vahicle-setup}
    \vspace{-2em}
\end{figure}

\begin{figure}[h!]
    \begin{minipage}{0.49\textwidth}
        \centering
         \includegraphics[width=.49\textwidth]{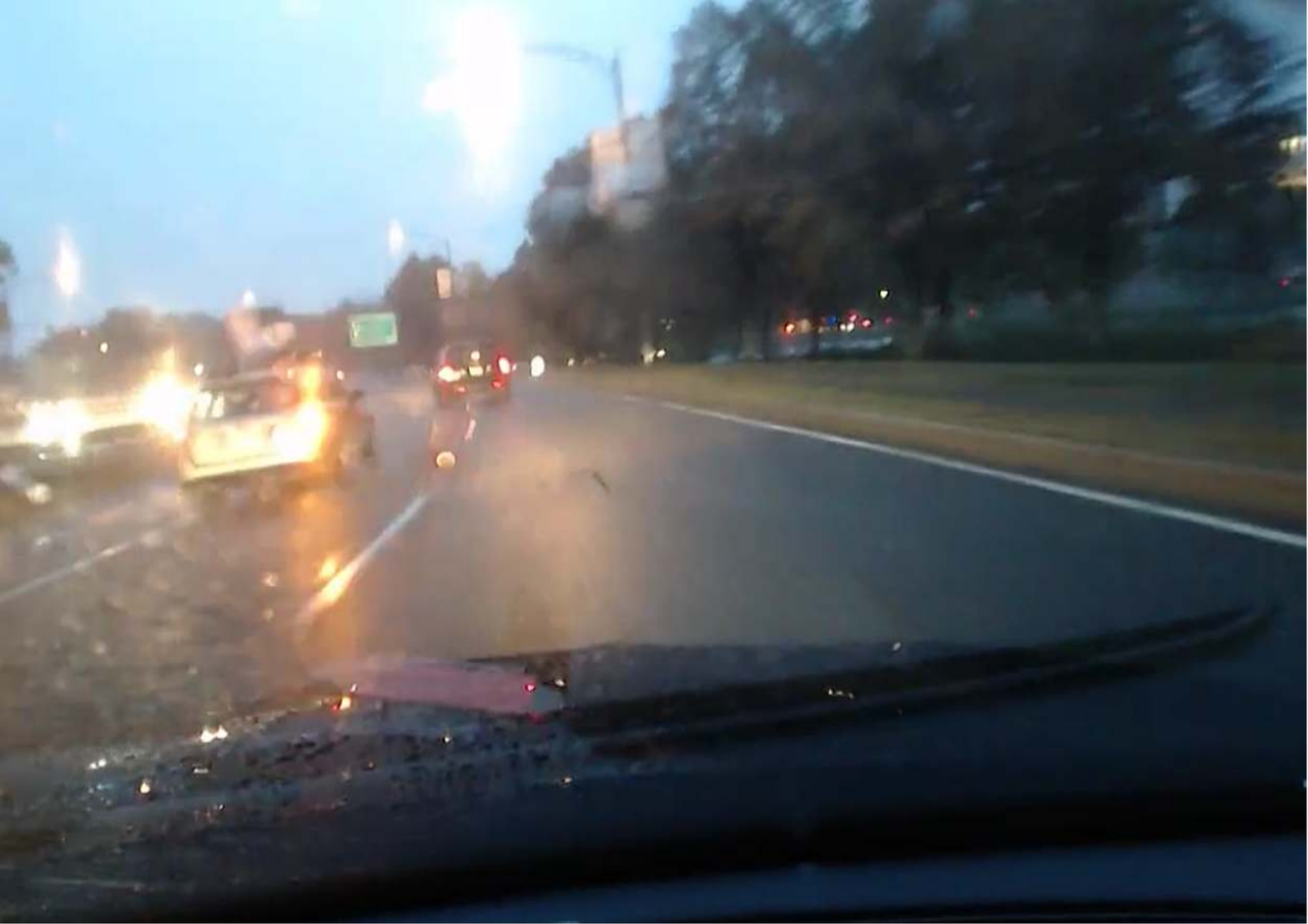}
         \includegraphics[width=.49\textwidth]{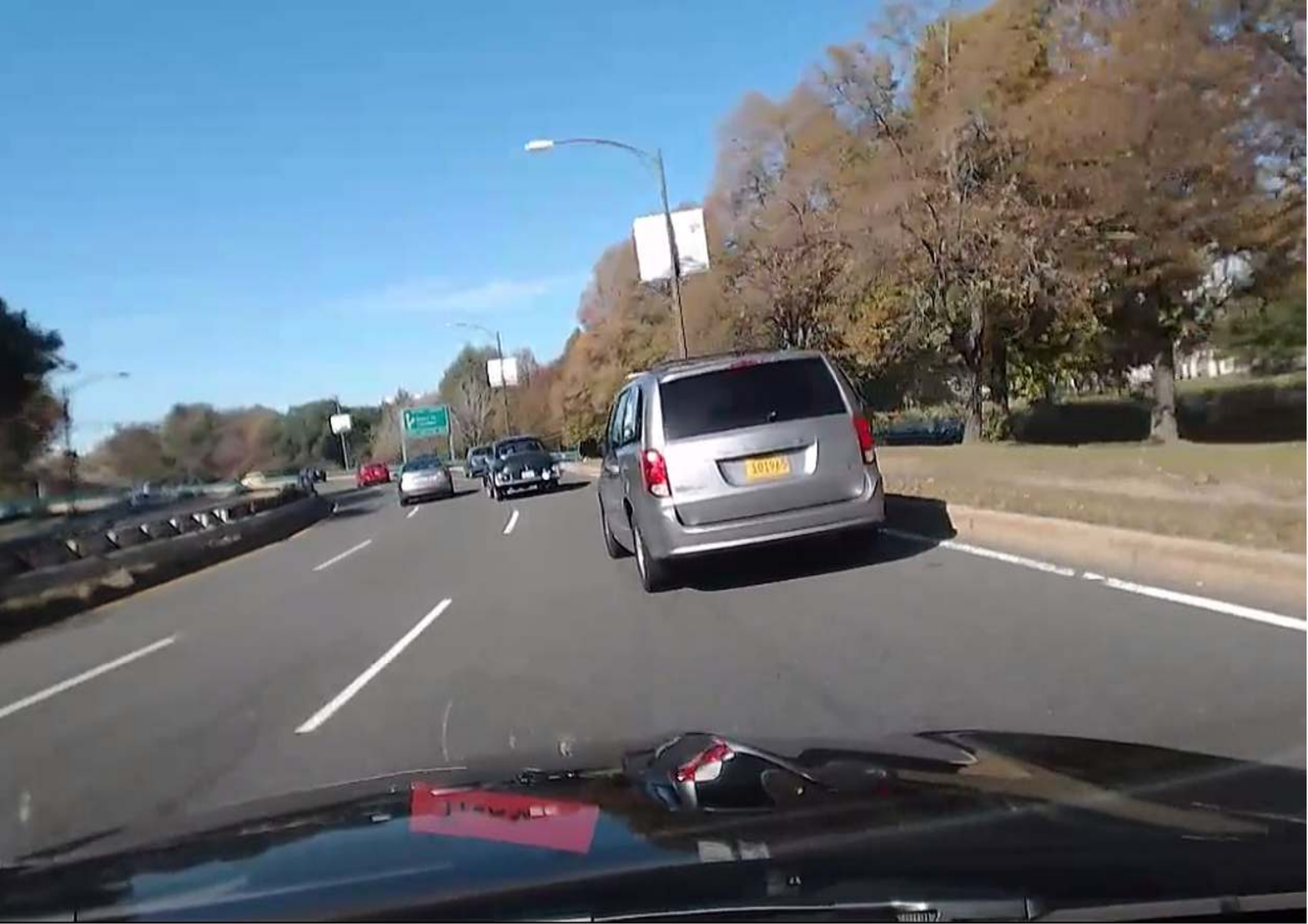}
        \caption[Snapshots from the video]%
        {Snapshots from the video of the forward roadway showing the same GPS location for a `wet' trip 1 (left) and a
          `dry' trip 1 (right). \footnote{A video of these trips is available at: \weblink{http://lexfridman.com/wetroad}}}
        \label{fig:same-road}
    \end{minipage}
    \vspace{-2em}
\end{figure}

\begin{figure}[h!]
  \centering
    \includegraphics[width=.24\textwidth]{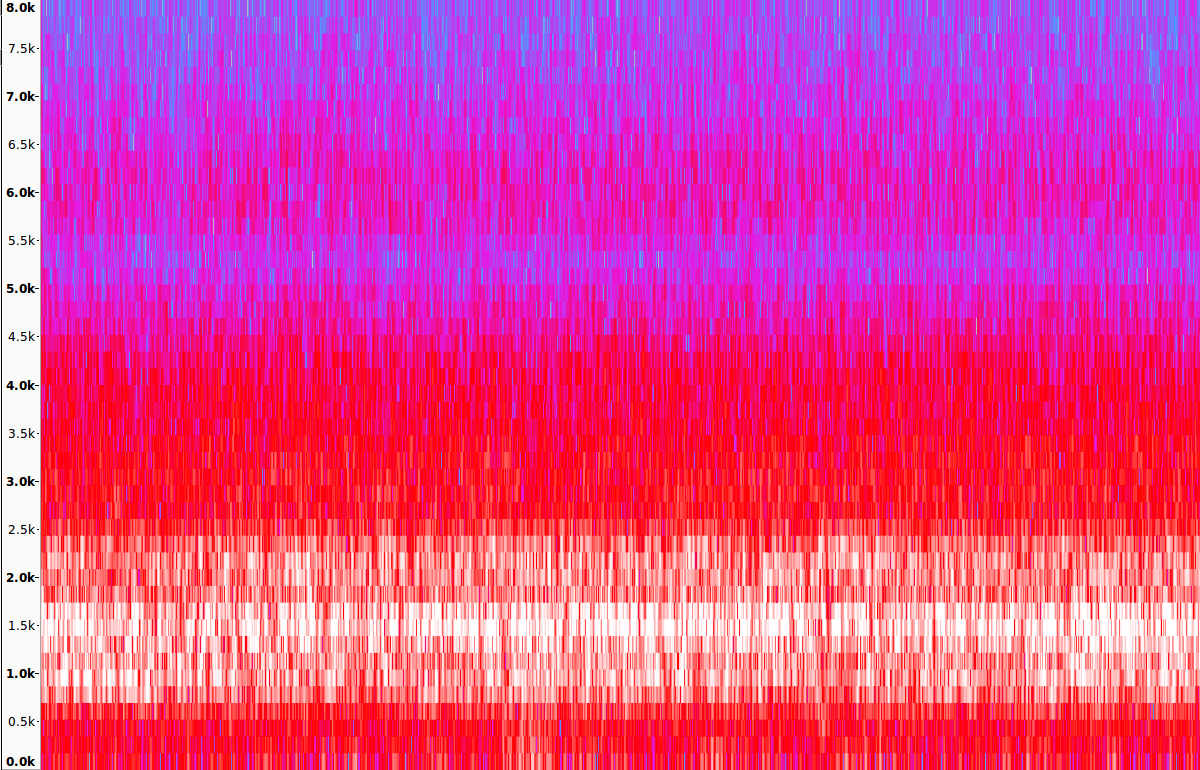}
    \includegraphics[width=.24\textwidth]{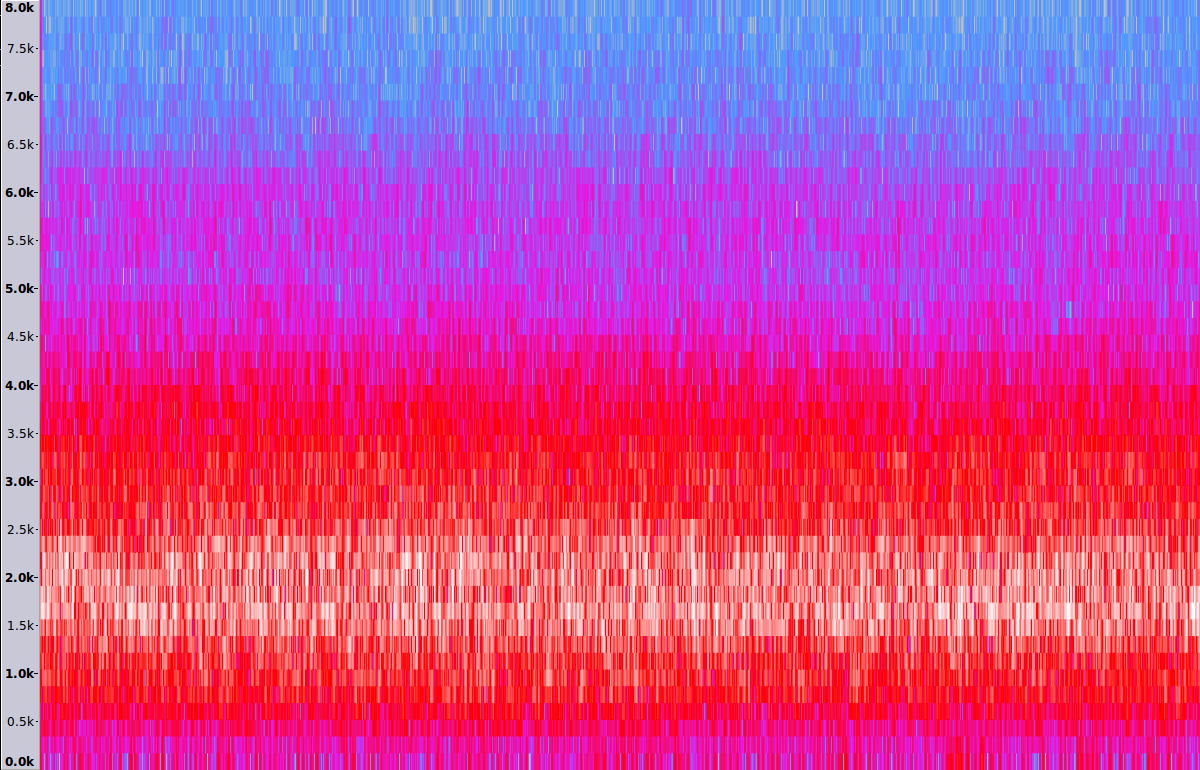}
    \caption{Spectrograms for the wet trip 2 (left) and dry trip 2 (right) from the same route segment at the speed of approximately 20\,mph.}
    \label{fig:spectrograms}
\end{figure}

For data collection purposes, we instrumented a 2014 Mercedes CLA with an inexpensive shotgun microphone behind the rear tire, as shown in \figref{vahicle-setup}. The gain level of the microphone and its distance from the tire were kept the same for the entire data collection process. Three different routes were selected. For each route, we drove the same exact path once during the rain (or immediately after) and another time when the road surface was completely dry, as shown in \figref{same-road}. We provide spectrograms in \figref{spectrograms} for wet and dry road segments of the same route that highlight the difference in frequency response. The duration and length of trips ranged from 14\,min to 30\,min and 6.1\,mi to 9.0\,mi, respectively. The summary of the dataset is presented in \tabref{data-summary}.

\begin{center}
    \begin{figure}[h!]
        \includegraphics[width=0.5\textwidth]{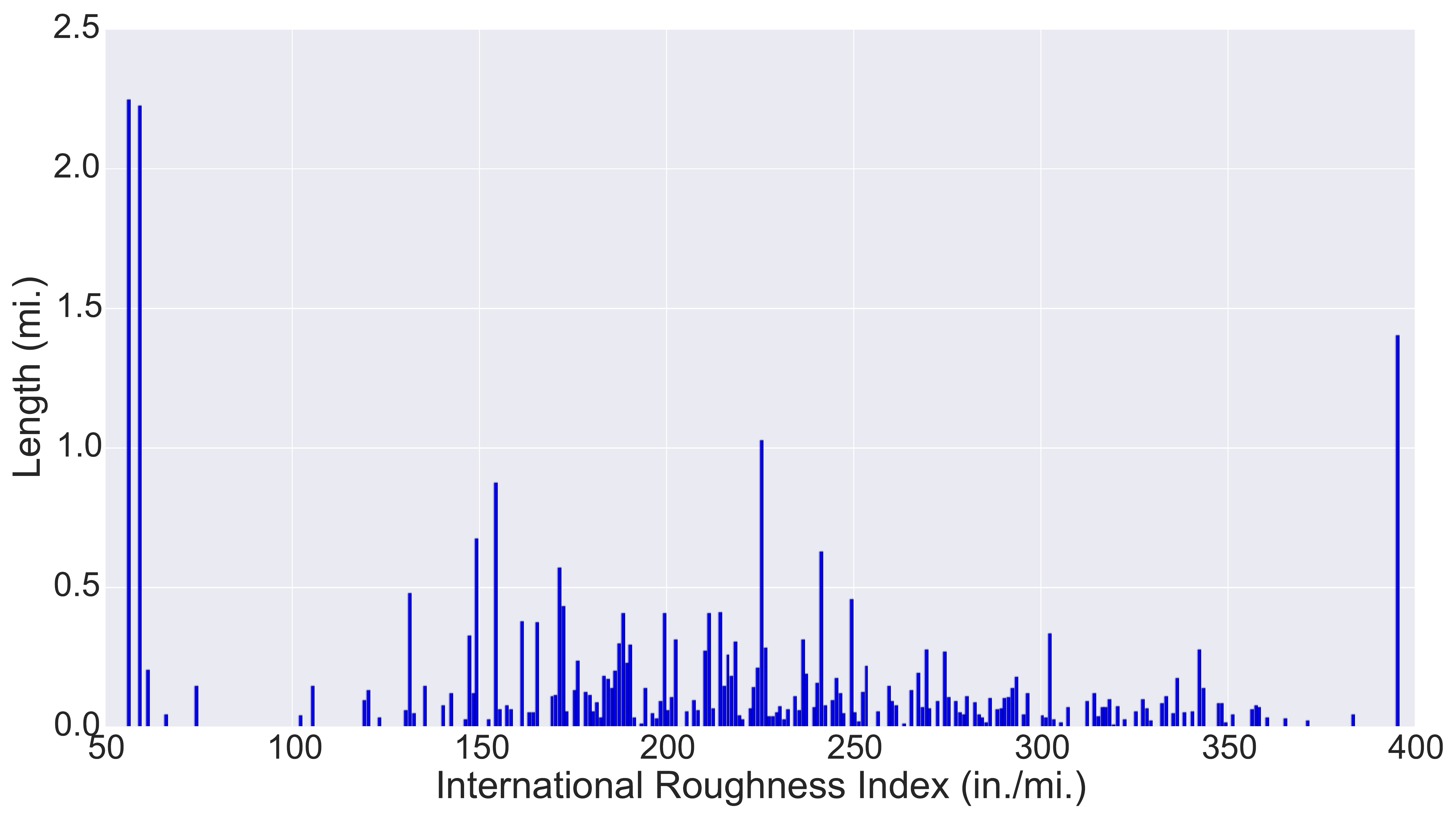}
        \caption{
            \label{fig:histogram-iri}
            Histogram or IRI distribution throughout collected data.
        }
        \vspace{-2em}
    \end{figure}
\end{center}

\begin{center}
    \begin{tabular}{ | l | l l l l|}
    \hline
    \textbf{Trip} & \textbf{Time} & \textbf{Distance} & \textbf{Avg Speed} & \textbf{Avg IRI}\\
    \hline
    wet 1 & 26\,min & 9.0\,mi & 7.4\,mph & 267\,in/mi\\
    wet 2 & 16\,min & 6.4\,mi & 9.4\,mph & 189\,in/mi\\
    wet 3 & 14\,min & 6.1\,mi & 13.5\,mph & 142\,in/mi\\
    dry 1 & 30\,min & 9.0\,mi & 9.6\,mph & 267\,in/mi\\
    dry 2 & 14\,min & 6.4\,mi & 9.1\,mph & 189\,in/mi\\
    dry 3 & 18\,min & 6.1\,mi & 9.3\,mph & 142\,in/mi\\
    \hline
    \end{tabular}
    \captionof{table}{Statistics of the collected data for six trips: time, distance, average speed and average IRI.} \label{tab:data-summary}
\end{center}
The data collection was carried out in Cambridge and the Greater Boston area with different speeds, traffic conditions and pavement roughness. The latter is measured with the International Roughness Index (IRI) which represents pavement quality \cite{paterson1986international}. A histogram of IRI values for the collected dataset is presented in \figref{histogram-iri}, wherein the unit of measurement is in inches per mile (in/mi). Our dataset contains values from 25\,in/mi to 1400\,in/mi, but in \figref{histogram-iri}, values over 400\,in/mi are aggregated into a single bin. According to the Massachusetts Department of Transportation (MassDOT) Road Inventory, the route we traveled is a combination of surface-treated road and bituminous concrete road \cite{massdot2015}.

\subsection{Features}\label{sec:audio-features}
Our aim was to model the whole spectrum along with the first order differences and then select a subset of features that discriminates our classes the best. We extracted Auditory Spectral Features (ASF)\cite{Marchi14-MLP}, that were computed by applying the short-time Fourier transform (STFT) using a frame size 30\,ms and a frame step of 10\,ms. Furthermore, each STFT power spectrogram has been converted to the Mel-Frequency scale using 26 triangular filters obtaining the Mel spectrograms $M^{30}(n, m)$. To match the human perception of loudness, a logarithmic representation has been chosen:
\begin{equation} M_{log}^{30}(n,m)=log(M^{30}(n,m)+1.0). \end{equation} 
In addition, the positive first order differences $D^{30}(n, m)$ were calculated from each Mel spectrogram as follows:
\begin{equation} D^{30}(n,m)=M_{log}^{30}(n,m)-M_{log}^{30}(n-1,m). \end{equation} 
The frame energy has also been included as a feature which resulted in a total of 54 features \cite{marchi2015non}. To foster reproducibility, we use the opensource software toolkits: (a) openSMILE -- for extracting features from the audio, and (b) Weka 3 -- for feature evaluation with Information Gain (IG) and Correlation-based Feature Selection (CFS) to reduce the dimension of the feature space \cite{eyben2013recent,hall2009weka}.

The IG feature evalaution is an univariate filter that calculates the worth of a feature by measuring the IG with respect to the class, it measures individual feature value but neglects redundancy \cite{karegowda2010comparative,hall2003benchmarking}. The output is a list of ranked features of which we selected best $5n$ features, where $n\in[1..10]$ and the whole feature set for comparison.

The CFS subset evaluation is a multivaraite filter that seeks for subsets of features that are highly correlated with the class while having low intercorrelation \cite{hall2009weka,Hall1998,karegowda2010comparative}. We used the BestFirst search algorithm in a forward search mode (-D 1) and a threshold of 5 non-improving nodes (-N 5) for consideration before terminating search. The CFS subset evaluation returned a list of 5 features.

\subsection{Classifier}\label{sec:audio-classifier}
In this work, we used a deep learning approach with initialized nets -- LSTM and bi-directional LSTM (BLSTM) RNN architectures which in contrast to other RNNs do not suffer from the problem of vanishing gradients \cite{hochreiter1997long}. The BLSTM is an extension of the LSTM architecture that allows for an additional forward pass if a look-ahead buffer may be used, which has been proven successful in many applications \cite{graves2005framewise}. 

In addition, we evaluated different parameters, such as the layout of LSTM and BLSTM hidden layers (54-54-54, 54-30-54, 156-256-156, 216-216-216, 216-316-216 neurons in the three hidden layers) and learning rates (1e-4, 1e-5, 1e-6). Initially, we chose deep architecture with three hidden layers of the same size as input vectors (54), before we ranked features and reduced its dimensionality. In the next step we investigated effectiveness of internal feature compression and augmentation of hidden layers to model more information. We used feed forward output layer with a logistic activation function and sum of squared error as objective function. The experiments were carried out with the CURRENNT toolkit \cite{Currennt2014}.

\section{Results}\label{sec:results}
\tabref{results} shows the evaluation results in an ascending order for the best 20 features that were selected with IG (IG-20), as described in \secref{audio-features} and trained with LSTM-RNNs. We present only the worst three and the best three results for RNNs, whereas other experiments were left out from the table. For every combination of parameters we conducted cross-validation on all three folds from \tabref{data-summary}. I.e., we leave out wet/dry 3 at a time for training with wet/dry 1 and testing with wet/dry 2, and run six experiments in total. Furthermore, an average UAR was computed for results obtained from all speeds including vehicle stationary mode. The best result with an UAR of 93.2\,\% was achieved with BLSTM network layout 216-216-216 and learning rate $1e^{-5}$. 

Additionally, we compared our results with the state-of-the-art approach of \cite{alonso2014board} that uses zero-norm minimization (L0) to select four most promising features (L0-4) from 125 ms audio bins of 1/3 octave bands (5000\,Hz, 1600\,Hz, 630\,Hz and 200\,Hz frequency bands). We trained SVMs with Sequential Minimal Optimization (SMO) on our dataset and found a C parameter of $1e^{-3}$ to give the best UAR of 67.4\,\%. Furthermore, experiments with SVMs and IG-20 feature set were carried out and gave the best UAR of 78.8\,\%.

\begin{center}
  \begin{tabular}{ |lll||ll| }
  \hline
    \rowcolor{gray!20}
    Network & Feature set & C\,(1e$^{-n}$) & UAR\,(\%) & DIFF\\
    \hline
    SVM & Z0-4 & 3 & 67.4 & +4.2\\
    SVM & IG-20 & 3 & 78.8 & +3.0\\
    \hline
    \hline
    \rowcolor{gray!20}
    Network & Layout & LR\,(1e$^{-n}$) & UAR\,(\%) & DIFF\\
    \hline
    LSTM & 216-216-216 & 4 & 66.3 & -20.3\\
    BLSTM & 216-316-216 & 4 & 76.1 & -10.5\\
    LSTM & 156-256-156 & 4 & 78.0 & -8.6\\
    \vdots & \vdots & \vdots & \vdots & \vdots\\
    BLSTM & 216-316-216 & 5 & 92.6 & +6.0\\
    LSTM & 216-216-216 & 5 & 92.6 & +6.0\\
    \textbf{BLSTM} & \textbf{216-216-216} & \textbf{5} & \textbf{93.2} & \textbf{+6.6}\\
    \hline
  \end{tabular}
  \captionof{table}{Comparison of results (upper) that were obtained by applying state-of-the-art approach of Alonso \ea, and (lower) our approach with RNNs, both trained and tested on our dataset. The column LR is an abbreviation for Learning Rate, and the column DIFF is an abbreviation for difference from the mean UAR.} \label{tab:results}
\end{center}

The mean UAR value for experiments with LSTM-RNNs is 86.6\,\% and the standard deviation equals 6.4. The mean UAR of all experiments with BLSTM network is 87.0\,\%, while the mean UAR for experiments with LSTM network is 86.0\,\%. The best mean UAR for experiments with learning rate $1e^{-5}$ amounts to 90.8\,\%, while the worst performing learning rate $1e^{-4}$ achieves only 78.8\,\%. 

Two out of three wet trips have significantly higher number of false predictions (1) at the beginning, where vehicle tires were dry before getting wetted from the surface, and (2) at the end of the trip, when the vehicle entered a parking lot with relatively dry road surface. 

\begin{figure}[h!]
  \centering
  \begin{subfigure}[b]{\columnwidth}\includegraphics[width=\columnwidth]{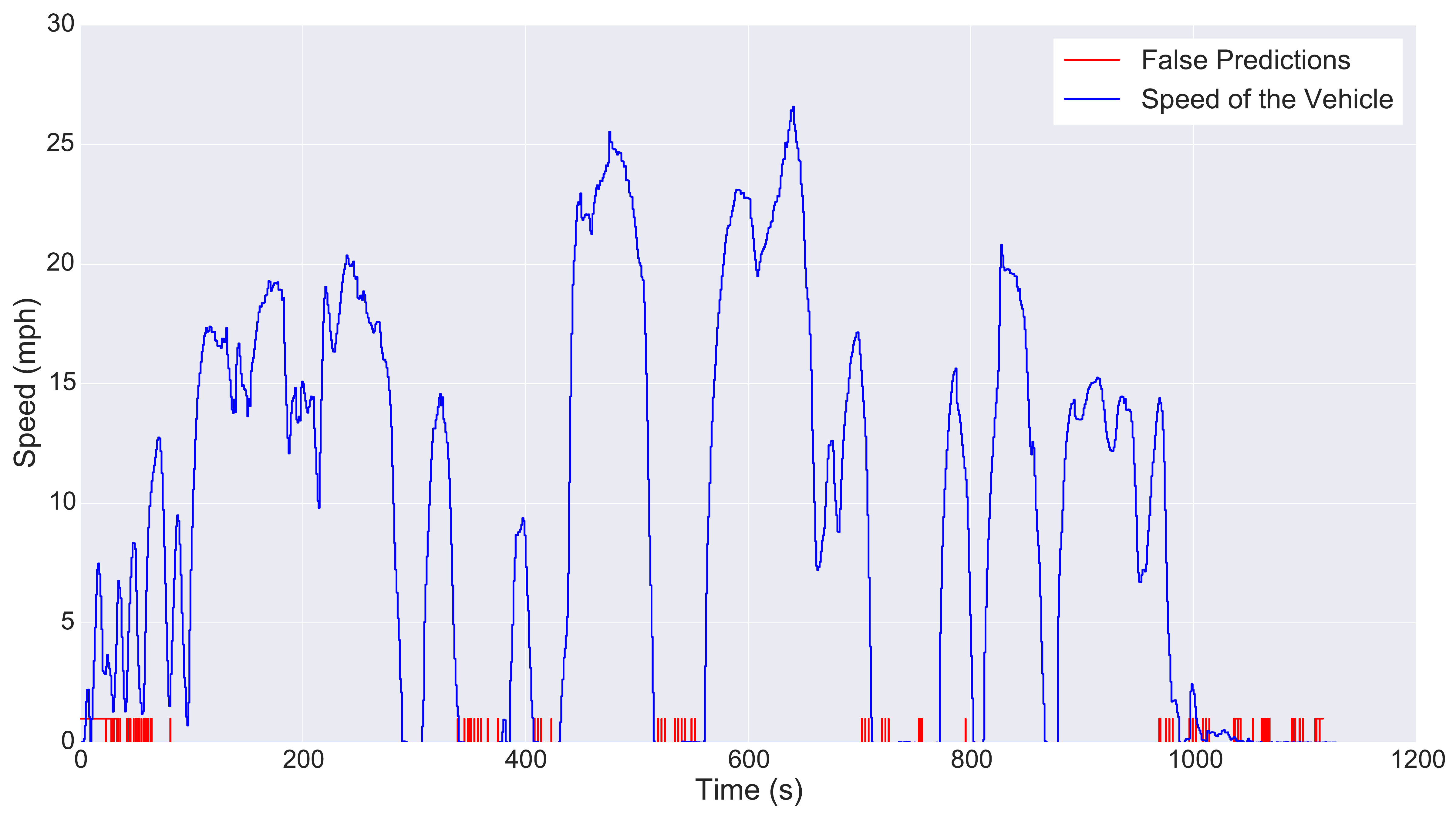}
    \caption{An 18\,min long wet trip showing speed and false predictions.}
    \label{fig:wet-fp}
  \end{subfigure}\\\vspace{0.1in}
  \begin{subfigure}[b]{\columnwidth}\includegraphics[width=\columnwidth]{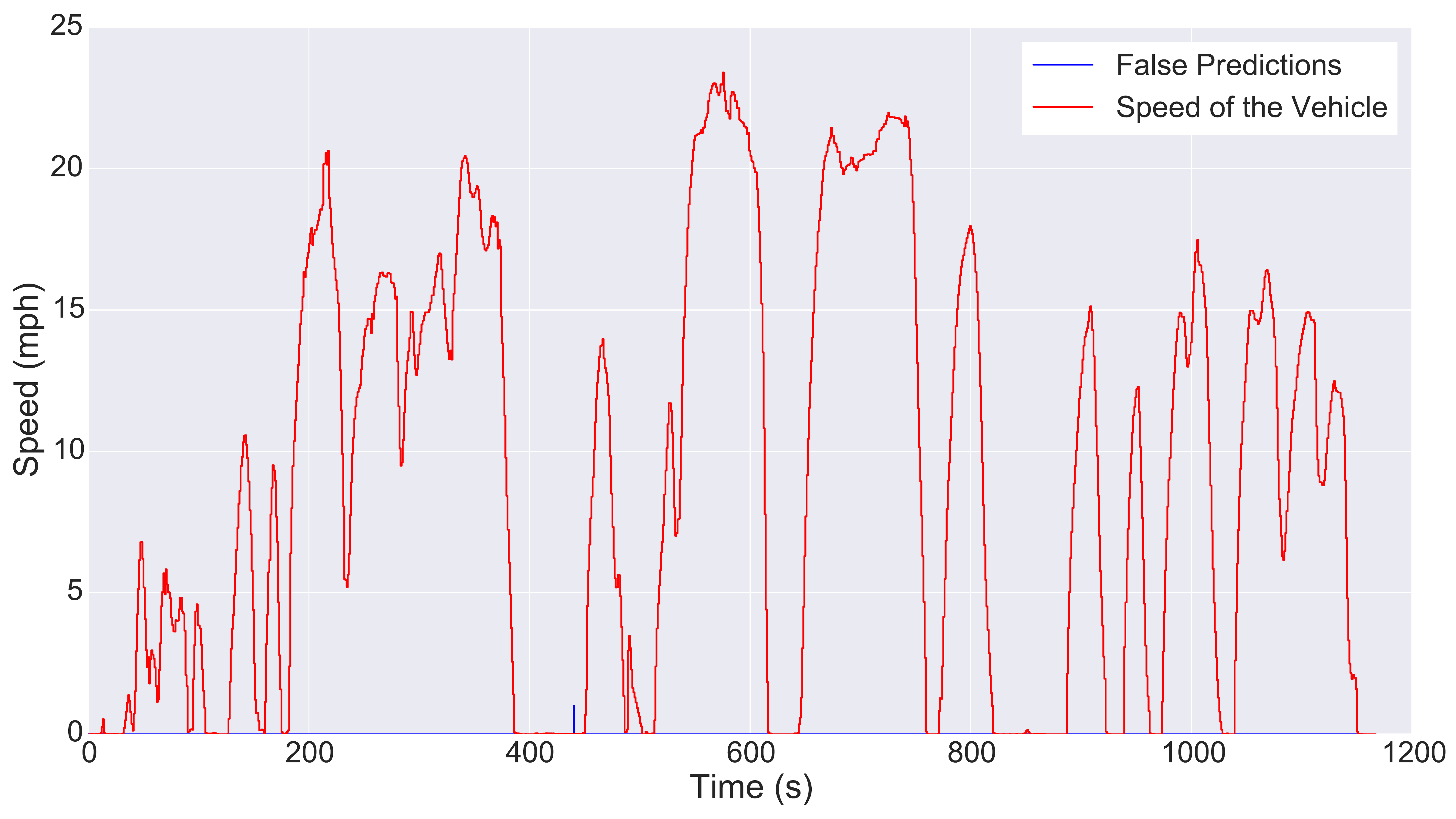}
    \caption{A 19\,min long dry trip showing speed and false predictions.}
    \label{fig:dry-fp}
  \end{subfigure}%
  \caption{Graphs for wet and dry road surfaces for route two that clarify the correlation between low speed and inaccurate predictions.}
  \label{fig:wet-dry}
  \vspace{-2em}
\end{figure} 

In \figref{wet-dry} we compare speed and false predictions of wet and dry trips for the same route that has similar properties, which are described in \secref{data-collection}. One can observe that all false predictions of wet trip 2 in \figref{wet-fp} occured below the speed of 2.9\,mph, whilst \figref{dry-fp} depicts a dry trip 2 and has only one false prediction when the vehicle is not moving. Therefore, discarding speeds below 2.9\,mph improves the UAR to 100\,\%. When we look only at speeds below 2.9\,mph and ignore everything above we are still able to attain 74.5\,\% UAR. The latter is possible only in presence of ambient sounds, as noises of vehicles that are driving by. 

\section{Conclusion}\label{sec:conclusion}
We proposed a deep learning approach based on LSTM-RNNs for detecting road wetness from audio of the tire-surface interaction and discriminating between wet and dry classes. This method is shown to be robust to vehicle speed, road type, and pavement quality on a dataset containing 785,826 bins of audio. It outperforms the state-of-the-art SVMs and achieves an outstanding performance on the road wetness detection task with an 93.2\,\% UAR for all vehicle speeds and the more challenging speeds being those below 2.9\,mph, including vehicle stationary mode. In future work, we will augment the feature set for estimating depth of water on the road surface and detecting hydroplaning conditions.

 %\section*{Acknowledgment}
 
\bibliographystyle{IEEEtran}
\bibliography{wet}

\begin{thebibliography}{10}
\providecommand{\url}[1]{#1}
\csname url@samestyle\endcsname
\providecommand{\newblock}{\relax}
\providecommand{\bibinfo}[2]{#2}
\providecommand{\BIBentrySTDinterwordspacing}{\spaceskip=0pt\relax}
\providecommand{\BIBentryALTinterwordstretchfactor}{4}
\providecommand{\BIBentryALTinterwordspacing}{\spaceskip=\fontdimen2\font plus
\BIBentryALTinterwordstretchfactor\fontdimen3\font minus
  \fontdimen4\font\relax}
\providecommand{\BIBforeignlanguage}[2]{{%
\expandafter\ifx\csname l@#1\endcsname\relax
\typeout{** WARNING: IEEEtran.bst: No hyphenation pattern has been}%
\typeout{** loaded for the language `#1'. Using the pattern for}%
\typeout{** the default language instead.}%
\else
\language=\csname l@#1\endcsname
\fi
#2}}
\providecommand{\BIBdecl}{\relax}
\BIBdecl

\bibitem{hamilton2012weather}
\BIBentryALTinterwordspacing
Booz-Allen-Hamilton, ``Ten-year averages from 2002 to 2012 based on nhtsa
  data,'' \emph{US Department of Transportation - Federal Highway
  Administration}, 2012. [Online]. Available:
  \url{www.ops.fhwa.dot.gov/weather}
\BIBentrySTDinterwordspacing

\bibitem{andrey2003weather}
J.~Andrey, B.~Mills, M.~Leahy, and J.~Suggett, ``Weather as a chronic hazard
  for road transportation in canadian cities,'' \emph{Natural Hazards},
  vol.~28, no. 2-3, pp. 319--343, 2003.

\bibitem{andrey2001weather}
J.~Andrey, B.~Mills, and J.~Vandermolen, ``Weather information and road
  safety,'' \emph{Institute for Catastrophic Loss Reduction, Toronto, Ontario,
  Canada}, 2001.

\bibitem{mueller2015sensor}
M.~Mueller, ``Sensor sensibility: Advanced driver assistance systems,''
  \emph{Vision Zero International}, 2015.

\bibitem{alonso2014board}
J.~Alonso, J.~L{\'o}pez, I.~Pav{\'o}n, M.~Recuero, C.~Asensio, G.~Arcas, and
  A.~Bravo, ``On-board wet road surface identification using tyre/road noise
  and support vector machines,'' \emph{Applied Acoustics}, vol.~76, pp.
  407--415, 2014.

\bibitem{graves2009novel}
A.~Graves, M.~Liwicki, S.~Fern{\'a}ndez, R.~Bertolami, H.~Bunke, and
  J.~Schmidhuber, ``A novel connectionist system for unconstrained handwriting
  recognition,'' \emph{Pattern Analysis and Machine Intelligence, IEEE
  Transactions on}, vol.~31, no.~5, pp. 855--868, 2009.

\bibitem{mayer2008system}
H.~Mayer, F.~Gomez, D.~Wierstra, I.~Nagy, A.~Knoll, and J.~Schmidhuber, ``A
  system for robotic heart surgery that learns to tie knots using recurrent
  neural networks,'' \emph{Advanced Robotics}, vol.~22, no. 13-14, pp.
  1521--1537, 2008.

\bibitem{graves2005framewise}
A.~Graves and J.~Schmidhuber, ``Framewise phoneme classification with
  bidirectional lstm and other neural network architectures,'' \emph{Neural
  Networks}, vol.~18, no.~5, pp. 602--610, 2005.

\bibitem{wollmer2008abandoning}
M.~W{\"o}llmer, F.~Eyben, S.~Reiter, B.~Schuller, C.~Cox, E.~Douglas-Cowie, and
  R.~Cowie, ``Abandoning emotion classes-towards continuous emotion recognition
  with modelling of long-range dependencies.'' in \emph{INTERSPEECH}, vol.
  2008, 2008, pp. 597--600.

\bibitem{xu2014experimental}
Y.~Xu, J.~Du, L.-R. Dai, and C.-H. Lee, ``An experimental study on speech
  enhancement based on deep neural networks,'' \emph{Signal Processing Letters,
  IEEE}, vol.~21, no.~1, pp. 65--68, 2014.

\bibitem{weninger2011audio}
F.~Weninger and B.~Schuller, ``Audio recognition in the wild: Static and
  dynamic classification on a real-world database of animal vocalizations,'' in
  \emph{acoustics, speech and signal processing (ICASSP), 2011 IEEE
  international conference on}.\hskip 1em plus 0.5em minus 0.4em\relax IEEE,
  2011, pp. 337--340.

\bibitem{eck2002finding}
D.~Eck and J.~Schmidhuber, ``Finding temporal structure in music: Blues
  improvisation with lstm recurrent networks,'' in \emph{Neural Networks for
  Signal Processing, 2002. Proceedings of the 2002 12th IEEE Workshop
  on}.\hskip 1em plus 0.5em minus 0.4em\relax IEEE, 2002, pp. 747--756.

\bibitem{Marchi14-MLP}
E.~Marchi, G.~Ferroni, F.~Eyben, L.~Gabrielli, S.~Squartini, and B.~Schuller,
  ``{Multi-resolution Linear Prediction Based Features for Audio Onset
  Detection with Bidirectional LSTM Neural Networks},'' in \emph{{Proceedings
  39th IEEE International Conference on Acoustics, Speech, and Signal
  Processing, ICASSP 2014}}, IEEE.\hskip 1em plus 0.5em minus 0.4em\relax
  Florence, Italy: IEEE, May 2014, pp. 2183--2187, (acceptance rate: 50\,\%,
  IF* 1.16 (2010)).

\bibitem{horita2012efficient}
Y.~Horita, S.~Kawai, T.~Furukane, and K.~Shibata, ``Efficient distinction of
  road surface conditions using surveillance camera images in night time,'' in
  \emph{Image Processing (ICIP), 2012 19th IEEE International Conference
  on}.\hskip 1em plus 0.5em minus 0.4em\relax IEEE, 2012, pp. 485--488.

\bibitem{yamada2003study}
M.~Yamada, T.~Oshima, K.~Ueda, I.~Horiba, and S.~Yamamoto, ``A study of the
  road surface condition detection technique for deployment on a vehicle,''
  \emph{JSAE review}, vol.~24, no.~2, pp. 183--188, 2003.

\bibitem{jokela2009road}
M.~Jokela, M.~Kutila, and L.~Le, ``Road condition monitoring system based on a
  stereo camera,'' in \emph{Intelligent Computer Communication and Processing,
  2009. ICCP 2009. IEEE 5th International Conference on}.\hskip 1em plus 0.5em
  minus 0.4em\relax IEEE, 2009, pp. 423--428.

\bibitem{amthor2015road}
M.~Amthor, B.~Hartmann, and J.~Denzler, ``Road condition estimation based on
  spatio-temporal reflection models,'' pp. 3--15, 2015.

\bibitem{jonsson2015road}
P.~Jonsson, J.~Casselgren, and B.~Thornberg, ``Road surface status
  classification using spectral analysis of nir camera images,'' \emph{Sensors
  Journal, IEEE}, vol.~15, no.~3, pp. 1641--1656, 2015.

\bibitem{viikari2009road}
V.~V. Viikari, T.~Varpula, and M.~Kantanen, ``Road-condition recognition using
  24-ghz automotive radar,'' \emph{Intelligent Transportation Systems, IEEE
  Transactions on}, vol.~10, no.~4, pp. 639--648, 2009.

\bibitem{iwao1996study}
K.~Iwao and I.~Yamazaki, ``A study on the mechanism of tire/road noise,''
  \emph{JSAE review}, vol.~17, no.~2, pp. 139--144, 1996.

\bibitem{bao2009acoustical}
M.~Bao, C.~Zheng, X.~Li, J.~Yang, and J.~Tian, ``Acoustical vehicle detection
  based on bispectral entropy,'' \emph{Signal Processing Letters, IEEE},
  vol.~16, no.~5, pp. 378--381, 2009.

\bibitem{birken2012voters}
R.~Birken, G.~Schirner, and M.~Wang, ``Voters: design of a mobile multi-modal
  multi-sensor system,'' in \emph{Proceedings of the Sixth International
  Workshop on Knowledge Discovery from Sensor Data}.\hskip 1em plus 0.5em minus
  0.4em\relax ACM, 2012, pp. 8--15.

\bibitem{fridman2015automated}
L.~Fridman, D.~E. Brown, W.~Angell, I.~Abdi{\'c}, B.~Reimer, and H.~Y. Noh,
  ``Automated synchronization of driving data using vibration and steering
  events,'' \emph{arXiv preprint arXiv:1510.06113}, 2015.

\bibitem{paterson1986international}
W.~D. Paterson, ``International roughness index: Relationship to other measures
  of roughness and riding quality,'' \emph{Transportation Research Record}, no.
  1084, 1986.

\bibitem{massdot2015}
\BIBentryALTinterwordspacing
MassDOT, ``Road inventory - massdot planning,'' 2015. [Online]. Available:
  \url{https://www.massdot.state.ma.us/planning/Main/MapsDataandReports/Data/GISData/RoadInventory.aspx}
\BIBentrySTDinterwordspacing

\bibitem{marchi2015non}
E.~Marchi, F.~Vesperini, F.~Weninger, F.~Eyben, S.~Squartini, and B.~Schuller,
  ``{Non-Linear Prediction with LSTM Recurrent Neural Networks for Acoustic
  Novelty Detection},'' in \emph{{Proceedings 2015 International Joint
  Conference on Neural Networks (IJCNN)}}, IEEE.\hskip 1em plus 0.5em minus
  0.4em\relax Killarney, Ireland: IEEE, July 2015, pp. 1--7.

\bibitem{eyben2013recent}
F.~Eyben, F.~Weninger, F.~Gross, and B.~Schuller, ``Recent developments in
  opensmile, the munich open-source multimedia feature extractor,'' in
  \emph{Proceedings of the 21st ACM international conference on
  Multimedia}.\hskip 1em plus 0.5em minus 0.4em\relax ACM, 2013, pp. 835--838.

\bibitem{hall2009weka}
M.~Hall, E.~Frank, G.~Holmes, B.~Pfahringer, P.~Reutemann, and I.~H. Witten,
  ``The weka data mining software: an update,'' \emph{ACM SIGKDD explorations
  newsletter}, vol.~11, no.~1, pp. 10--18, 2009.

\bibitem{karegowda2010comparative}
A.~G. Karegowda, A.~Manjunath, and M.~Jayaram, ``Comparative study of attribute
  selection using gain ratio and correlation based feature selection,''
  \emph{International Journal of Information Technology and Knowledge
  Management}, vol.~2, no.~2, pp. 271--277, 2010.

\bibitem{hall2003benchmarking}
M.~Hall, G.~Holmes \emph{et~al.}, ``Benchmarking attribute selection techniques
  for discrete class data mining,'' \emph{Knowledge and Data Engineering, IEEE
  Transactions on}, vol.~15, no.~6, pp. 1437--1447, 2003.

\bibitem{Hall1998}
M.~A. Hall, ``Correlation-based feature subset selection for machine
  learning,'' Ph.D. dissertation, University of Waikato, Hamilton, New Zealand,
  1998.

\bibitem{hochreiter1997long}
S.~Hochreiter and J.~Schmidhuber, ``Long short-term memory,'' \emph{Neural
  computation}, vol.~9, no.~8, pp. 1735--1780, 1997.

\bibitem{Currennt2014}
J.~Weninger, Felix~Bergmann and B.~Schuller, ``Introducing currennt - the
  munich open-source cuda recurrent neural network toolkit,'' \emph{Journal of
  Machine Learning Research}, no.~16, pp. 547--551, 2014.

\end{thebibliography}

\end{document}